%% file: iclr2027_conference.tex
\documentclass{article}
\usepackage[T1]{fontenc}
\PassOptionsToPackage{table}{xcolor}
\usepackage{iclr2027_conference,times}

\usepackage{amsmath,amsfonts,amssymb}
\usepackage{graphicx}
\usepackage{placeins}
\usepackage{float}
\usepackage{wrapfig}
\usepackage{multirow,booktabs,array}
\usepackage{xcolor}
\usepackage{tikz}
\usetikzlibrary{arrows.meta,positioning,fit,calc,backgrounds}
\usepackage{xurl}
\usepackage{hyperref}

\definecolor{vslink}{HTML}{235B85}
\definecolor{vslinkbg}{HTML}{EDF4F8}
\newcommand{\vsiconcode}{\tikz[baseline=-0.45ex,x=0.55em,y=0.55em,line width=0.6pt]{%
  \draw (0.55,0.05)--(0.05,0.5)--(0.55,0.95);
  \draw (1.25,0.05)--(1.75,0.5)--(1.25,0.95);
  \draw (1.05,0)--(0.75,1);}}
\newcommand{\vsiconbench}{\tikz[baseline=-0.45ex,x=0.55em,y=0.55em,line width=0.6pt]{%
  \draw[rounded corners=0.5pt] (0.1,0.05) rectangle (1.6,0.95);
  \draw (0.35,0.7)--(0.65,0.7) (0.35,0.5)--(1.3,0.5) (0.35,0.3)--(1.1,0.3);}}
\newcommand{\vsiconcount}{\tikz[baseline=-0.45ex,x=0.55em,y=0.55em,line width=0.6pt]{%
  \draw (0.1,0.05)--(0.1,0.95) (0.8,0.05)--(0.8,0.95) (1.5,0.05)--(1.5,0.95);
  \fill (0.1,0.25) circle (0.1) (0.8,0.7) circle (0.1) (1.5,0.4) circle (0.1);}}
\newcommand{\vsiconpath}{\tikz[baseline=-0.45ex,x=0.55em,y=0.55em,line width=0.6pt]{%
  \draw (0.2,0.2)--(0.75,0.8)--(1.45,0.35);
  \fill (0.2,0.2) circle (0.13) (0.75,0.8) circle (0.13) (1.45,0.35) circle (0.13);}}
\newcommand{\vsresource}[3]{%
  \href{#1}{\colorbox{vslinkbg}{\textcolor{vslink}{\footnotesize\bfseries\strut #2\hspace{0.32em}#3}}}%
}

\newenvironment{contributionlist}{%
  \begin{list}{\textbullet}{%
    \setlength{\leftmargin}{1.1em}%
    \setlength{\rightmargin}{0pt}%
    \setlength{\labelwidth}{0.5em}%
    \setlength{\labelsep}{0.6em}%
    \setlength{\itemindent}{0pt}%
    \setlength{\listparindent}{0pt}%
    \setlength{\topsep}{0.4\baselineskip}%
    \setlength{\itemsep}{0.25\baselineskip}%
    \setlength{\parsep}{0pt}%
  }%
}{\end{list}}

\title{Visual Scaffolding:\\Toward Native Multimodal Reasoning}
\author{\begin{minipage}{\dimexpr\textwidth-2\tabcolsep\relax}
\centering
\vspace*{0.85em}
\normalsize\bfseries Run Shao \quad Zhaoyang Zhang \quad Ziyu Li \quad Linrui Xu \quad Ji Qi\\[0.45em]
Xiu Su \quad Yichao Cao \quad Chao Tao \quad Haifeng Li\textsuperscript{*}\\[0.65em]
\normalfont Central South University\\[0.2em]
\footnotesize\textsuperscript{*}Corresponding author
\end{minipage}}
\hypersetup{
  hidelinks,
  pdftitle={Visual Scaffolding: Toward Native Multimodal Reasoning},
  pdfauthor={Run Shao, Zhaoyang Zhang, Ziyu Li, Linrui Xu, Ji Qi, Xiu Su, Yichao Cao, Chao Tao, Haifeng Li},
  pdfkeywords={Multimodal reasoning, vision-language models, chain-of-thought, visual grounding, representation learning}
}

\iclrfinalcopy
\begin{document}
\maketitle
\vspace{-0.8em}
\begin{center}
\setlength{\fboxsep}{4pt}%
\vsresource{https://github.com/GeoX-Lab/VisualScaffolding}{\vsiconcode}{Code \& Evaluation}\hspace{0.35em}%
\vsresource{https://huggingface.co/datasets/ShaoRun/nmr-bench}{\vsiconbench}{NMR-Bench}\hspace{0.35em}%
\vsresource{https://huggingface.co/ShaoRun/vscaffold-4b-burden-counting}{\vsiconcount}{Counting Model}\hspace{0.35em}%
\vsresource{https://huggingface.co/ShaoRun/vscaffold-4b-path-tracing}{\vsiconpath}{Path Model}
\end{center}
\vspace{-0.25em}

\begin{abstract}
The success of chain-of-thought (CoT) in complex reasoning can be attributed to a textual scaffolding effect: explicit intermediate results can be examined, reused, and composed to support subsequent reasoning, much as scaffolding supports a building under construction. However, mainstream multimodal LLMs lack an equivalent scaffold for visual distinctions that are difficult to express in concise verbal descriptions. We therefore ask: \emph{Can native visual representations become first-class citizens alongside language, providing a visual scaffold for reasoning?} We define \emph{native multimodal reasoning} as reasoning in which native visual representations and language jointly carry reusable intermediate states. As a concrete implementation, the \emph{visual scaffolding} mechanism we proposed lets the MLLM autonomously initiate a retrieval action at any reasoning step. A lightweight retriever selects native visual features, which are then incorporated directly into the evolving CoT and remain available to subsequent steps. To evaluate this mechanism under controlled visual demands, we introduce NMR-Bench, comprising Burden Counting and Path Tracing. With 88M added parameters (approximately 2\%) on Qwen3-VL-4B-Thinking, our model achieves 66.43\% and 72.50\% accuracy, respectively, surpassing the evaluated GPT-5.5 and Gemini 3.1 Pro baselines. Controlled interventions show that changing selected visual content affects task accuracy and can redirect subsequent visual selection even when intervening text is held fixed. These findings show that native visual states can function as intermediate reasoning material and point toward reusable scaffolds for additional modalities.
\end{abstract}

\suppressfloats[t]
\begin{figure}[t]
\centering
\includegraphics[width=\textwidth]{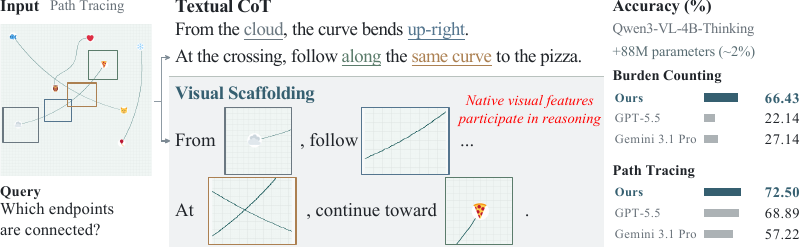}
\caption{\textbf{From textual CoT to visual scaffolding.} Textual CoT describes spatial relations in words. Visual scaffolding inserts native visual features as reusable intermediate material. Colored boxes link source regions to illustrative feature views. Right: NMR-Bench accuracy with a 4B backbone and 88M added parameters. Traces are schematic.}
\label{fig:visual-scaffolding}
\end{figure}

\section{Introduction}
Chain-of-thought (CoT) has become a standard component of frontier LLMs, supporting advances across complex reasoning tasks, including mathematics and coding \citep{wei2022,openai_o1,deepseek_r1} and agentic problem solving \citep{react}. We attribute this success to a \emph{textual scaffolding effect}: intermediate results remain available for examination, reuse, and composition. Much as scaffolding supports a building under construction, a textual trace lets each reasoning step build on earlier results.

This view is consistent with accounts of external representation in human problem solving. Persistent artifacts keep information available for inspection and manipulation \citep{zhang1994,kirsh2010}, while language stabilizes ideas for revision \citep{clark1998}. In LLM reasoning, scratchpads expose intermediate computation \citep{nye2021}, and least-to-most prompting reuses earlier subproblem answers to solve later ones \citep{least_to_most}.

However, mainstream MLLMs still express intermediate decisions mainly in words \citep{zhang2024,qwen3vl}. The input image remains accessible, but concise descriptions may not preserve the visual identity or spatial relation established at an earlier step \citep{blink,jin2026}. This raises a broader question: \textbf{\emph{Can native visual representations become first-class citizens alongside language, providing a visual scaffold for reasoning?}}

We define \emph{native multimodal reasoning} (NMR) as reasoning in which language and native visual representations jointly carry reusable intermediate states. In this form of reasoning, visual content enters the evolving CoT itself and remains available to later steps. It can therefore contribute beyond references expressed through textual coordinates or structured tool calls \citep{grit,vigorl,hu2024,pixelreasoner}.

Our \emph{visual scaffolding} mechanism provides a concrete implementation. The MLLM decides when to request visual information by generating a special token, \texttt{[ref]}, during CoT. A lightweight retriever then selects native visual features, which are adapted and inserted into the sequence before generation resumes. Earlier selected features remain in context as later reasoning unfolds, as illustrated in Figure~\ref{fig:visual-scaffolding}.

We aggregate reasoning context on the retriever's query side and encode spatial information on its key side. Dual-resolution encoding retains the backbone's standard image representation and supplies a finer feature bank for retrieval.

We introduce \emph{NMR-Bench} to evaluate this mechanism under controlled visual demands. \emph{Burden Counting} tests the ability to maintain instance identity during enumeration, while \emph{Path Tracing} tests spatial continuity through competing local connections. Procedural generation provides exact answers and spatial annotations, allowing us to relate task outcomes to the visual evidence selected during reasoning.

With 88M added parameters, our Qwen3-VL-4B-Thinking model reaches 66.43\% accuracy on Counting and 72.50\% on Path, exceeding the evaluated GPT-5.5 and Gemini 3.1 Pro baselines on both tasks. Even when intervening words are held fixed, changing selected visual content can redirect later retrieval.

Our contributions are as follows:
\begin{contributionlist}
\item We define NMR as reasoning in which native visual representations and language jointly carry reusable intermediate states, and propose visual scaffolding as a concrete implementation.
\item We introduce NMR-Bench to evaluate sustained visual demands and show that visual scaffolding surpasses the evaluated frontier MLLMs in overall accuracy on both tasks.
\item Controlled interventions and evidence-selection analyses show how native visual content affects subsequent reasoning and retrieval, suggesting a path toward CoT with additional native modalities.
\end{contributionlist}

\section{Related Work}
\textbf{Intermediate representations.}
CoT and scratchpads make intermediate computation available as text \citep{wei2022,nye2021}. Self-consistency and Tree of Thoughts build on textual traces by sampling or searching over them \citep{self_consistency,tree_of_thoughts}. Multimodal-CoT grounds textual rationales in images \citep{zhang2024}. Coconut, MVoT, and COVT explore continuous hidden states, image visualizations, and predicted visual tokens \citep{coconut,mvot,covt}. Visual scaffolding instead reuses features of the input image within the reasoning trace.

\textbf{Spatial and executable interfaces.}
VISPROG and ViperGPT organize visual reasoning as executable programs \citep{visprog,vipergpt}. Visual CoT, GRIT, and ViGoRL direct attention through region, box, or point interfaces \citep{viscot,grit,vigorl}. These interfaces specify where to look. Visual scaffolding places the selected native features into the sequence used by later steps.

\textbf{Visual tools and artifacts.}
Other approaches create or operate on visual artifacts. Visual Sketchpad creates sketches, and CogCoM manipulates images \citep{hu2024,cogcom}. o3, DeepEyes, and Pixel Reasoner use visual operations \citep{openai_o3,deepeyes,pixelreasoner}. Thyme executes code \citep{thyme}, while PyVision-RL and Mini-o3 extend tool interaction and visual search \citep{pyvisionrl,minio3}. Our mechanism selects source-linked content from the encoded input.

\section{Visual Scaffolding Mechanism}
\label{sec:method}
\subsection{Overview: Bringing Visual Representations into CoT}
\label{sec:scaffold-overview}
An MLLM receives an image and a question, then develops its CoT primarily through text generation. Our visual scaffolding mechanism lets the model add native image representations to that same reasoning sequence when it needs visual evidence. The model decides when to request this evidence by generating a special token, \texttt{[ref]}, alongside ordinary text tokens.

\begin{figure}[!htbp]
\centering
\includegraphics[width=\linewidth]{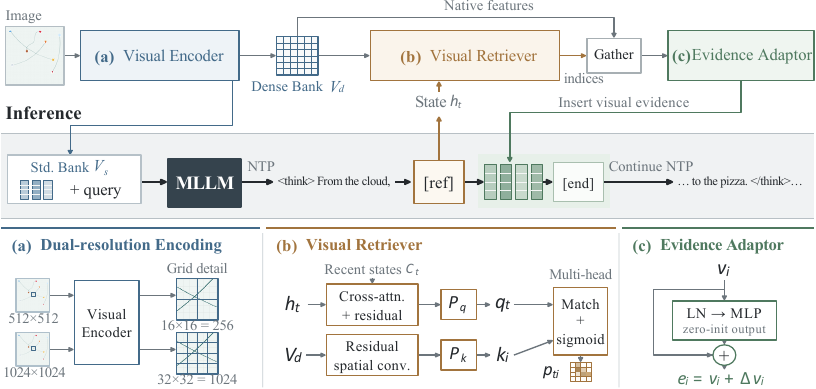}
\caption{\textbf{Visual scaffolding mechanism.} The MLLM generates text until \texttt{[ref]} initiates a visual request. The Visual Retriever selects native features from the dense visual bank, and the Evidence Adaptor inserts them into CoT before generation resumes. The standard visual bank supplies the original image input. Panels detail (a) dual-resolution encoding, (b) retrieval conditioned on reasoning context and local spatial structure, and (c) residual evidence adaptation.}
\label{fig:scaffolding-mechanism}
\end{figure}

The decoder state at \texttt{[ref]} represents the current reasoning need. A lightweight Visual Retriever uses that state to select features from the encoded image, and an Evidence Adaptor prepares the selected features for insertion into CoT. After processing the visual block and its closing token, \texttt{[end]}, the model resumes next-token prediction (NTP). It can issue another request as its reasoning develops. A trace thus alternates between generated text and visual blocks:
\begin{equation}
T_0\;\Vert\;[\texttt{[ref]},E_1,\texttt{[end]}]\;\Vert\;T_1\;\Vert\;[\texttt{[ref]},E_2,\texttt{[end]}]\;\Vert\;\cdots,
\label{eq:scaffold-trace}
\end{equation}
Here $T_j$ is generated text, $E_j$ is a selected visual block, and $\Vert$ denotes concatenation. Both the original image input and the inserted visual blocks remain in the decoder's causal context. Later steps can therefore use a visual detail selected earlier, alongside the text already written. Figure~\ref{fig:scaffolding-mechanism} illustrates this flow.

For a request to be useful, the available features must distinguish nearby image elements. We retain the backbone's standard image input and construct a finer feature bank for retrieval. Section~\ref{sec:scaffold-retrieval} explains how that bank is built and how the current reasoning state selects from it.

\subsection{Selecting Visual Support for the Current Step}
\label{sec:scaffold-retrieval}
\textbf{Dual-resolution encoding.}
Retrieval requires candidate features fine enough to distinguish nearby objects or path segments. MLLMs commonly merge visual-encoder outputs to limit the number of tokens passed to the language model. In Qwen3-VL, $16\times16$-pixel encoder patches are merged in $2\times2$ groups \citep{qwen3vl}. For a $512\times512$ input, this produces a $16\times16$ grid with a nominal 32-pixel stride. A merged feature may cover several small objects or competing path segments.

We obtain finer retrieval candidates while retaining the backbone's ordinary image input and its encoder--merger architecture. The \emph{same} Visual Encoder processes the image twice: the original view provides the standard visual bank $V_s$ for the input prefix, and a view enlarged twofold provides the dense visual bank $V_d=\{v_i\}_{i=1}^{N}$, with $v_i\in\mathbb{R}^d$, for retrieval. With a $512\times512$ source image, the enlarged $1024\times1024$ view yields a $32\times32$ grid: 1024 selectable features at a nominal 16-pixel stride in source coordinates. This gives the retriever four times as many candidate locations without enlarging the original image prefix or adding another visual encoder. Formally, $V_s=\mathcal{E}(I)$ and $V_d=\mathcal{E}(U_2(I))$, where $\mathcal{E}$ includes the shared encoder and spatial merger, and $U_2$ doubles both image dimensions. These grid sizes assume size-preserving preprocessing. The dense bank and its retrieval keys are computed once per image.

\textbf{Reasoning-conditioned queries.}
Each request should retrieve evidence relevant to the question the model is resolving at that step. The decoder state $h_t$ at \texttt{[ref]} captures the immediate need, while information about the relevant object or path may have appeared earlier in the trace. The retriever therefore uses cross-attention \citep{vaswani2017} to combine the current state with up to 256 preceding states $C_t$, including text and previously inserted visual evidence:
\begin{equation}
q_t=P_q\!\left(h_t+\operatorname{CrossAttn}(h_t,C_t)\right).
\label{eq:scaffold-query}
\end{equation}
The projection MLP $P_q$ maps this context-aware state into retrieval space. $C_t$ excludes the current request. The 256-state window applies only to the retriever; the decoder retains its full causal history.

\textbf{Spatially encoded keys.}
The query is matched against image locations. To represent the local structure around each candidate, a spatial encoder processes neighboring cells in the dense bank before key projection, without changing grid resolution:
\begin{equation}
S(V_d)=\operatorname{LN}\!\left(V_d+f_{\mathrm{sp}}(V_d)\right),\qquad
k_i=P_k\!\left(S(V_d)_i\right).
\label{eq:scaffold-keys}
\end{equation}
The spatial encoder $f_{\mathrm{sp}}$ uses channel projections and two residual blocks of $3\times3$ convolutions. The resulting \emph{keys} contain local context, while the \emph{values} inserted into CoT remain the original native features $v_i$. This lets spatial structure guide selection while preserving the selected visual content. Appendix~\ref{app:architecture} gives the dimensions.

\textbf{Selecting a set of cells.}
A single visual step may need more than one cell: an object can span several candidates, and a path segment may continue across neighboring locations. The retriever therefore assigns each cell an independent sigmoid score. The score combines $H=16$ heads of width $d_h=128$ with learned shared weights $\alpha=\operatorname{softmax}(\omega)$:
\begin{equation}
 p_{ti}=\sigma\!\left(\sum_{a=1}^{H}\alpha_a\frac{\langle q_t^{(a)},k_i^{(a)}\rangle}{\sqrt{d_h}}\right).
\label{eq:scaffold-scores}
\end{equation}
Cells scoring at least 0.5 are selected. If the set exceeds the inference cap $K_{\mathrm{inf}}$, the retriever keeps the highest-scoring cells; if no cell reaches the threshold, it keeps the highest-scoring one. The selected set $\mathcal{S}_t$ is placed in grid order to form a stable sequence of $K_t=|\mathcal{S}_t|$ evidence features. The cap controls the size of each inserted block, while later reasoning steps can initiate further requests.

\subsection{Making Selected Content Reusable}
\label{sec:scaffold-integration}
The selected features come from the visual encoder, but they now need to serve as intermediate material inside a text-generating decoder. The Evidence Adaptor learns this interface through a residual transformation that preserves feature width:
\begin{equation}
\phi(v_i)=v_i+W_2\operatorname{GELU}(W_1\operatorname{LN}(v_i)+b_1)+b_2,
\qquad E_t=[\phi(v_i)]_{i\in\mathcal{S}_t}.
\label{eq:scaffold-adapter}
\end{equation}
The output layer is initialized to zero, so the adaptor begins as an identity map and then learns how to present selected features to the decoder. At inference, the decoder processes $E_t$ and \texttt{[end]} before continuing text generation. The inserted features occupy positions in the autoregressive sequence, while the original image prefix retains the backbone's positional encoding. Subsequent predictions can attend to the visual block through causal attention.

The same process also links successive visual requests. Decoder states conditioned on earlier evidence can enter the retriever's context $C_t$ at a later \texttt{[ref]}. An inserted visual detail can therefore affect both the words generated next and the image regions selected later.

\subsection{Learning to Select and Use Evidence}
\label{sec:scaffold-learning}
\begin{figure}[!htbp]
\centering
\includegraphics[width=\linewidth]{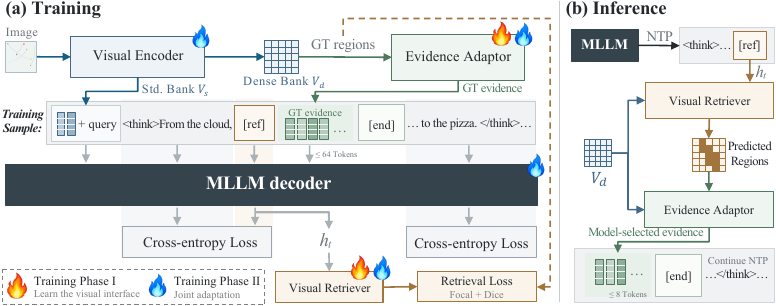}
\caption{\textbf{Learning to select and use visual evidence.} (a) During training, GT-selected features enter the reasoning sequence. Next-token prediction supervises text and request tokens, while focal and Dice losses supervise feature selection. Phase I emphasizes the new interface; Phase II jointly adapts it with the backbone. (b) During inference, the model initiates each request and selects the features inserted before generation resumes. Training allows up to 64 evidence tokens per request; the main evaluation caps predicted selections at 8.}
\label{fig:scaffolding-training}
\end{figure}

\textbf{Ground-truth evidence injection.}
The model must learn when to request evidence, which features to select, and how to continue reasoning after receiving them. Each training example supplies an image, a question, a reasoning trace, request positions, and ground-truth (GT) patch sets. At an annotated request, adapted GT-selected features are inserted while the trace is teacher-forced. The decoder thus learns to continue from the intended visual support, while supervision on retrieval scores teaches the retriever to select that support (Figure~\ref{fig:scaffolding-training}). At inference, request timing and feature selection use model predictions rather than GT patches.

\textbf{Joint supervision.}
The decoder generates both reasoning text and request tokens, so next-token prediction trains request timing together with the continuation after visual evidence. For a mixed text-and-visual prefix $Z_{<t}$, the objective is
\begin{equation}
\mathcal{L}_{\mathrm{ntp}}=-\frac{1}{|\mathcal{M}|}\sum_{t\in\mathcal{M}}\log P_\theta(y_t\mid Z_{<t}).
\label{eq:scaffold-ntp}
\end{equation}
The supervised set $\mathcal{M}$ contains reasoning text, request tokens, final answers, and assistant termination. Image-prefix positions, inserted evidence, and reference-closing positions are masked as prediction targets. Predictions after an evidence block can still propagate gradients through the visual features they use.

Relevant cells usually occupy only a small part of the dense bank. We therefore use focal loss, with positive and negative cells normalized separately, to supervise retrieval scores \citep{lin_focal2017}. This gives annotated cells sufficient weight against the larger set of background cells. For request $r$, let $g_{ri}$ denote binary GT membership, $p_{ri}$ the predicted score, and $\mathcal{P}_r,\mathcal{N}_r$ the positive and negative sets:
\begin{equation}
\mathcal{L}_{\mathrm{focal}}^{(r)}=
-\frac{\sum_{i\in\mathcal{P}_r}(1-p_{ri})^\gamma\log(p_{ri}+\epsilon)}{|\mathcal{P}_r|}
-\frac{\sum_{i\in\mathcal{N}_r}p_{ri}^\gamma\log(1-p_{ri}+\epsilon)}{|\mathcal{N}_r|}.
\label{eq:scaffold-focal}
\end{equation}
We use $\gamma=2$ and $\epsilon=10^{-7}$; an empty class contributes zero. Cellwise scores do not directly express overlap with the full annotated region, so we also use a soft Dice loss \citep{dice1945}:
\begin{equation}
\mathcal{L}_{\mathrm{Dice}}^{(r)}=1-\frac{2\sum_i p_{ri}g_{ri}+1}{\sum_i p_{ri}+\sum_i g_{ri}+1}.
\label{eq:scaffold-dice}
\end{equation}
The Dice calculation spans the candidate grid and uses a smoothing constant of 1. For $R$ annotated requests, the total objective combines next-token prediction with retrieval loss averaged over annotated requests:
\begin{equation}
\mathcal{L}=\mathcal{L}_{\mathrm{ntp}}+\frac{\lambda_R}{R}\sum_{r=1}^{R}\left(\mathcal{L}_{\mathrm{focal}}^{(r)}+\lambda_D\mathcal{L}_{\mathrm{Dice}}^{(r)}\right).
\label{eq:scaffold-objective}
\end{equation}
We set $\lambda_R=1$ for both tasks and $\lambda_D=1$ for Path and 2 for Counting. Examples without annotated requests contribute zero retrieval loss. Together, these objectives train the model to select relevant features and continue reasoning from the inserted evidence.

\textbf{Two-phase learning and compact inference.}
The Visual Retriever and Evidence Adaptor are newly added modules, while the backbone already encodes images and generates text. Phase I gives the new interface a larger learning rate and keeps small, nonzero learning rates for the language and vision backbone. Phase II jointly adapts selection, insertion, and continuation. All parameter groups remain trainable (Appendix~\ref{app:optimization}).

Training allows up to 64 evidence tokens per request to retain each annotated GT patch set. At inference, predicted selections are capped at $K_{\mathrm{inf}}=8$ to keep individual visual blocks compact. A complete request uses $K_t+2$ tokens, including its opening and closing markers, within the shared text-and-evidence generation budget. The added interface contains 88.15M parameters, or 1.99\% of Qwen3-VL-4B-Thinking.

\section{NMR-Bench}
\label{sec:nmr_bench}
NMR-Bench evaluates reasoning with reusable visual states under two demands that persist across steps: maintaining the identity of individual instances and preserving spatial continuity. Burden Counting and Path Tracing instantiate these demands in procedurally generated scenes. Exact answers and spatial annotations allow us to examine both task outcomes and the visual support selected during reasoning. Representative inputs and traces appear in the case study in Section~\ref{sec:exp_cases}.

\textbf{Burden Counting.}
Counting in cluttered scenes requires a model to distinguish similar objects and retain those distinctions while maintaining a tally. Objects vary in category, size, tilt, occlusion, and surrounding distractors. The benchmark contains seven question forms. Five require integer counts under category or attribute constraints, including joint size-and-tilt conditions. The other two compare counts between size groups or identify the most frequent category. The test set contains 420 images: 20 for each combination of seven forms and three difficulty levels. Difficulty is specified within each form.

\textbf{Path Tracing.}
Tracing a curve through intersections requires a model to preserve its identity while choosing among locally competing continuations. Each image contains intersecting, non-branching curves whose endpoints are marked with letters, numbers, icons, or shapes. Easy, Medium, and Hard scenes contain 3, 4--5, and 6--7 curves, respectively. The test set contains 360 images, with 120 at each difficulty level and 90 for each endpoint-marker type.

\textbf{Training and metrics.}
The training sets contain 32,000 Counting and 40,000 Path examples, each with a reasoning trace and ground-truth (GT) patch sets. At evaluation time, the model receives only the image and question. Let $y_i,\hat y_i$ be canonical and predicted answers, assigning invalid predictions $\bot$. Counting answers are integers or label--value tuples. For its five integer-answer forms, $\mathcal{V}$ contains valid numerical predictions $\hat c_i$ with targets $c_i$. Path answers are sets of unordered endpoint pairs $\mathcal{P}_i,\widehat{\mathcal{P}}_i$. We report
\begin{equation}
\mathrm{Acc}=\frac{\sum_{i=1}^{n}\mathbf{1}[\hat y_i=y_i]}{n},\quad
\mathrm{MAE}=\frac{\sum_{i\in\mathcal{V}}|\hat c_i-c_i|}{|\mathcal{V}|},\quad
\mathrm{CRA}=\frac{1}{n_P}\sum_{i=1}^{n_P}\frac{|\widehat{\mathcal{P}}_i\cap\mathcal{P}_i|}{|\mathcal{P}_i|}.
\label{eq:bench-metrics}
\end{equation}
Here $n$ is the task's test size and $n_P$ the Path test size. Acc. requires the complete predicted answer to match the target, and invalid outputs count as incorrect. MAE averages count error over valid numerical predictions, using $|\mathcal{V}|$ as its denominator. Connection Recovery Accuracy (CRA) measures the fraction of true connections recovered in each image and averages that fraction across the Path test set. For an invalid output or unknown endpoint label, $\widehat{\mathcal{P}}_i=\varnothing$. Acc. and CRA are percentages; MAE is in count units. Appendix~\ref{app:benchmark-details} specifies scoring rules and data distributions.

\section{Experiments}
\label{sec:experiments}

\subsection{Setup}
\label{sec:exp_setup}

\begin{table}[!t]
\centering
\caption{\textbf{NMR-Bench results.} VScaffold-4B is our visual-scaffolding model. Acc. and CRA are percentages. MAE uses integer-answer tasks. $\Delta_{\mathrm{NoRef}}=M_{\mathrm{ours}}-M_{\mathrm{NoRef}}$ reports signed gains over the trigger-only No-Ref control trained on the same data without inserted visual features. Higher Acc./CRA and lower MAE are better. Bold marks column-best values. Training and comparison settings are in Appendices~\ref{app:optimization} and~\ref{app:main-results}.}
\label{tab:nmr-bench-main}
\fontsize{8}{9.2}\selectfont
\setlength{\tabcolsep}{0.8pt}
\renewcommand{\arraystretch}{1.08}
\begin{tabular}{@{}l*{16}{r}@{}}
\toprule
\multirow{3}{*}{\textbf{Model}} & \multicolumn{8}{c}{\textbf{Burden Counting}} & \multicolumn{8}{c}{\textbf{Path Tracing}}\\
\cmidrule(lr){2-9}\cmidrule(lr){10-17}
& \multicolumn{2}{c}{Overall} & \multicolumn{2}{c}{Easy} & \multicolumn{2}{c}{Medium} & \multicolumn{2}{c}{Hard}
& \multicolumn{2}{c}{Overall} & \multicolumn{2}{c}{Easy} & \multicolumn{2}{c}{Medium} & \multicolumn{2}{c}{Hard}\\
\cmidrule(lr){2-3}\cmidrule(lr){4-5}\cmidrule(lr){6-7}\cmidrule(lr){8-9}\cmidrule(lr){10-11}\cmidrule(lr){12-13}\cmidrule(lr){14-15}\cmidrule(lr){16-17}
& Acc. & MAE & Acc. & MAE & Acc. & MAE & Acc. & MAE
& Acc. & CRA & Acc. & CRA & Acc. & CRA & Acc. & CRA\\
\midrule
\rowcolor[HTML]{F1F2F2}\multicolumn{17}{c}{\textit{Frontier Multimodal LLMs}}\\
MiniMax-M3 & 4.76 & 9.14 & 10.71 & 2.84 & 3.57 & 7.92 & 0.00 & 16.66 & 18.33 & 36.77 & 40.00 & 53.89 & 14.17 & 35.50 & 0.83 & 20.91 \\
Claude-Opus-4.8 & 6.43 & 9.38 & 14.29 & 3.03 & 2.86 & 8.27 & 2.14 & 16.85 & 33.06 & 54.09 & 66.67 & 76.11 & 29.17 & 52.71 & 3.33 & 33.45 \\
Kimi-K2.6 & 9.29 & 15.67 & 20.71 & 3.19 & 5.00 & 11.87 & 2.14 & 31.66 & 58.06 & 68.71 & 76.67 & 81.67 & 62.50 & 73.50 & 35.00 & 50.97 \\
Qwen-3.7-Plus & 18.81 & 17.52 & 37.14 & 1.84 & 11.43 & 20.57 & 7.86 & 30.14 & 44.44 & 63.33 & 73.33 & 80.00 & 44.17 & 61.29 & 15.83 & 48.69 \\
GPT-5.5 & 22.14 & 2.88 & 35.71 & 1.54 & 18.57 & 2.80 & 12.14 & 4.29 & 68.89 & 84.20 & 86.67 & 91.11 & 70.00 & 84.12 & \textbf{50.00} & \textbf{77.36} \\
Gemini-3.1-Pro & 27.14 & 3.25 & 48.57 & 1.01 & 22.14 & 2.84 & 10.71 & 5.86 & 57.22 & 77.73 & 85.00 & 90.00 & 56.67 & 76.92 & 30.00 & 66.27 \\
\rowcolor[HTML]{F1F2F2}\multicolumn{17}{c}{\textit{Coordinate-Mediated Reasoning Models}}\\
GRIT & 1.43 & 14.29 & 2.86 & 9.10 & 1.43 & 16.33 & 0.00 & 17.44 & 0.00 & 14.47 & 0.00 & 24.22 & 0.00 & 14.97 & 0.00 & 9.57 \\
ViGoRL & 0.71 & 14.36 & 2.14 & 6.18 & 0.00 & 12.80 & 0.00 & 24.11 & 0.28 & 11.06 & 0.83 & 21.39 & 0.00 & 12.17 & 0.00 & 5.53 \\
\rowcolor[HTML]{F1F2F2}\multicolumn{17}{c}{\textit{Multimodal Visual-Tool Agents}}\\
o3 & 6.90 & 6.34 & 12.86 & 2.92 & 5.00 & 6.78 & 2.86 & 9.35 & 10.28 & 22.25 & 22.50 & 36.94 & 8.33 & 25.47 & 0.00 & 13.24 \\
DeepEyes & 1.90 & 12.20 & 3.57 & 5.12 & 2.14 & 10.74 & 0.00 & 21.38 & 0.28 & 13.33 & 0.83 & 22.50 & 0.00 & 11.38 & 0.00 & 6.11 \\
Pixel Reasoner & 2.14 & 10.96 & 2.86 & 5.44 & 2.86 & 9.50 & 0.71 & 17.95 & 0.56 & 15.97 & 1.67 & 26.11 & 0.00 & 15.36 & 0.00 & 11.70 \\
Thyme & 3.57 & 9.90 & 7.14 & 3.75 & 2.14 & 8.22 & 1.43 & 17.73 & 0.83 & 16.51 & 2.50 & 28.89 & 0.00 & 16.48 & 0.00 & 10.80 \\
PyVision-RL & 1.90 & 13.55 & 4.29 & 5.12 & 1.43 & 11.92 & 0.00 & 23.67 & 0.83 & 10.83 & 2.50 & 22.22 & 0.00 & 8.43 & 0.00 & 7.20 \\
Mini-o3 & 2.38 & 42.77 & 5.00 & 22.52 & 2.14 & 32.95 & 0.00 & 73.57 & 0.56 & 12.38 & 1.67 & 20.56 & 0.00 & 15.17 & 0.00 & 6.68 \\
\rowcolor[HTML]{F1F2F2}\multicolumn{17}{c}{\textit{Based on Qwen3-VL-4B-Thinking}}\\
Qwen3-VL-4B-T & 2.86 & 10.49 & 5.71 & 3.85 & 2.14 & 9.18 & 0.71 & 18.43 & 4.72 & 18.45 & 13.33 & 29.72 & 0.83 & 16.25 & 0.00 & 9.38 \\
No-Ref-4B & 7.38 & 4.64 & 11.43 & 3.02 & 5.71 & 4.36 & 5.00 & 6.58 & 53.06 & 68.17 & 87.50 & 90.00 & 50.83 & 68.67 & 20.83 & 45.85 \\
\textbf{VScaffold-4B} & \textbf{66.43} & \textbf{0.50} & \textbf{82.86} & \textbf{0.22} & \textbf{67.14} & \textbf{0.56} & \textbf{49.29} & \textbf{0.71} & \textbf{72.50} & \textbf{86.98} & \textbf{94.17} & \textbf{95.83} & \textbf{82.50} & \textbf{91.29} & 40.83 & 73.83 \\
\rowcolor[HTML]{E8F1F5}$\Delta_{\mathrm{NoRef}}$ & 59.05 & -4.14 & 71.43 & -2.80 & 61.43 & -3.80 & 44.29 & -5.87 & 19.44 & 18.81 & 6.67 & 5.83 & 31.67 & 22.62 & 20.00 & 27.98 \\
\bottomrule
\end{tabular}
\end{table}

Our experiments address three questions. First, does visual scaffolding improve reasoning under sustained visual demands? Second, does the visual content inserted into CoT affect later answers and visual selections? Third, how does the model select evidence as the evidence budget and visual demands change?

For the first question, NMR-Bench compares VScaffold-4B with frontier MLLMs as general-purpose reference points, and with coordinate-mediated models and visual-tool agents as alternative ways to engage with an image during reasoning. The unmodified Qwen3-VL-4B-Thinking backbone establishes its starting performance. The No-Ref control uses the same task data and retains the \texttt{[ref]} trigger but does not retrieve or insert new visual features. Its comparison with VScaffold-4B measures the gain from visual scaffolding under this training setup.

\subsection{Main Results}
\label{sec:exp_same_backbone}\label{sec:exp_main}
VScaffold-4B achieves 66.43\% accuracy on Counting and 72.50\% on Path (Table~\ref{tab:nmr-bench-main}). These results exceed the strongest evaluated proprietary baseline for each task by 39.29 and 3.61 percentage points (pp), respectively. Compared with the trigger-only No-Ref control trained on the same task data, visual scaffolding improves accuracy by 59.05 pp on Counting and 19.44 pp on Path. The comparison shows the benefit of inserting native visual features for tasks requiring instance discrimination and spatial continuity.

Counting accuracy is 82.86\%, 67.14\%, and 49.29\% on Easy, Medium, and Hard, with the best Acc. and MAE at each level. On Medium Path cases, VScaffold-4B reaches 82.50\% exact-answer accuracy and recovers 91.29\% of connections. The count-error and connection-recovery results show that the gains extend beyond the exact-match metric.

\subsection{Case Study: Visual States in Use}
\label{sec:exp_cases}
\begin{figure}[!htbp]
\centering\includegraphics[width=\linewidth]{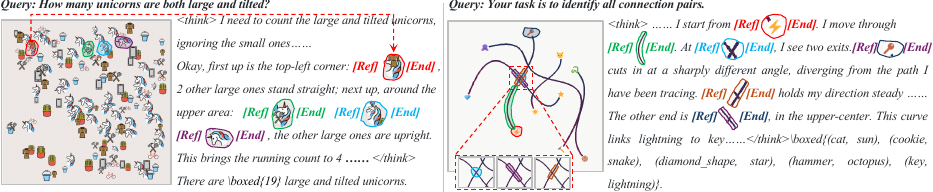}
\caption{\textbf{Visual intermediate states in two benchmark settings.} Counting interleaves a running tally with features selected from large, tilted unicorns (left). Path Tracing uses selected curve segments to recover endpoint connections (right). Outlines and insets indicate the source regions of the selected features. Traces are abbreviated and edited for presentation.}
\label{fig:benchmark-cases}
\end{figure}

Figure~\ref{fig:benchmark-cases} shows representative reasoning traces with selected visual evidence for both tasks.

\subsection{Visual Content Shapes Subsequent Reasoning}
\label{sec:exp_causal}
\begin{figure}[!htbp]
\centering\includegraphics[width=\linewidth]{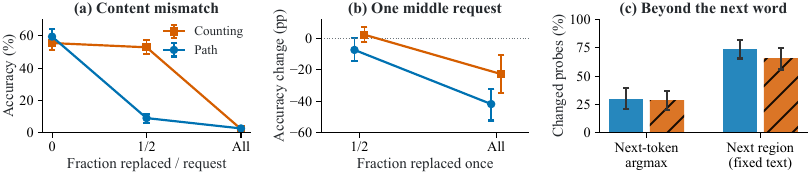}
\caption{\textbf{Effects of changing visual intermediate content.} (a) Replacing features at every request changes task accuracy. (b) Replacing one middle block measures the effect of a local intervention relative to unmodified replay. (c) Immediate next-token choices and subsequent visual selections capture different responses to a changed block. Intervening words are fixed only when comparing subsequent selections. Fractions indicate the share of features replaced; whiskers show 95\% bootstrap intervals. The evidence cap is 32.}
\label{fig:content-feedback}\label{tab:evidence_content}
\end{figure}

We test the role of inserted visual content by replacing selected features with features from other cells in the same image. The model weights and original image input remain fixed, as do the visual block's length, markers, and positions. Figure~\ref{fig:content-feedback} summarizes these tests; Appendix~\ref{app:evidence} specifies donor sampling and replay.

\textbf{Selected content supports task completion.}
Replacing every visual block completely lowers Counting accuracy from 55.48\% to 2.38\% and Path accuracy from 59.44\% to 2.78\%. Replacing half the features lowers Path accuracy by 50.28 pp, while mean output lengths remain similar (1,655 versus 1,697 tokens) and the invalid-output rate rises by only 1.11 pp. The Path accuracy change therefore cannot be attributed to answer formatting alone. At a single middle block, complete replacement has a larger effect than half replacement by 25.00 pp on Counting and 34.38 pp on Path.

\textbf{Visual feedback under fixed wording.}
When the intervening words are held fixed, changing one visual block changes the next selection in 55/84 Counting examples and 71/96 Path examples. The immediate next-token choice changes in 24/84 and 29/96 examples, respectively. In a separate free-continuation check, 33/96 Path examples produce the same nonempty words through the next request but select different patches. Together, these observations show that altered visual content can redirect a subsequent visual request without a preceding change in emitted words. Additional donor and feature-norm controls appear in Appendix~\ref{app:evidence}.

\subsection{Selectivity of Visual Intermediate States}
\label{sec:exp_reference_policy}\label{sec:exp_efficiency}
Using fixed checkpoints, we compare evidence caps $K\in\{8,16,32\}$ to test how the size of each block affects accuracy.

\begin{wrapfigure}[13]{r}{.54\linewidth}
\centering
\includegraphics[width=.96\linewidth]{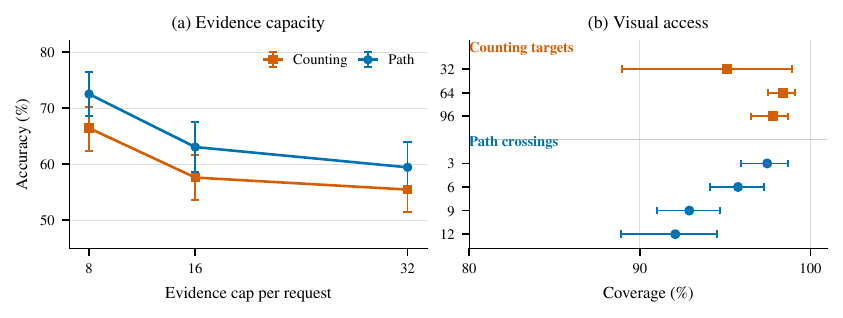}
\caption{\textbf{Selective visual support.} (a) Accuracy at evidence caps 8, 16, and 32. (b) Coverage at cap 32. Whiskers show 95\% bootstrap intervals; the panel (b) axis starts at 80\%.}
\label{fig:capacity-access}
\end{wrapfigure}

Using a cap of 8 instead of 32 improves Counting accuracy by 10.95 pp and Path accuracy by 13.06 pp (Figure~\ref{fig:capacity-access}(a)). To assess uncertainty in these differences, we resample test images while keeping each image's two predictions paired. The resulting 95\% bootstrap intervals for the accuracy gains are $[6.43,15.48]$ and $[8.61,17.50]$ pp. Mean evidence tokens per image fall from 204.3 to 171.8 on Counting and from 764.9 to 346.6 on Path. Because the cap applies to each request, the model can still add further evidence at later steps. Appendix~\ref{app:budget} reports evidence-token use at each cap.

\subsection{Visual Access Across Increasing Demands}
\label{sec:exp_burden}\label{sec:exp_difficulty}
In Counting, we vary the number of targets while holding 128 object locations fixed. In Path, we vary the number of crossings while retaining six curves. We measure \emph{coverage} as the fraction of annotated task-relevant pixels overlapped by the union of regions selected across the complete reasoning trace. Counting coverage is averaged over objects; Path coverage is measured over curve pixels. The annotation masks are used only after inference to calculate this diagnostic measure.

With 32, 64, and 96 targets, Counting reaches 95.14\%, 98.40\%, and 97.81\% coverage (Figure~\ref{fig:capacity-access}(b)). Path coverage remains above 92\% as crossings increase from 3 to 12. The selected regions thus continue to reach task-relevant image content across these controlled demands. Appendix~\ref{app:burden} gives the corresponding generator settings and coverage calculation.
\FloatBarrier

\section{Discussion and Conclusion}
Visual scaffolding changes what a CoT trace can preserve. Model-selected native features enter the causal sequence and remain accessible to later steps, so visual distinctions need not be expressed solely in words. Selected visual content can thus play the persistent role that intermediate text plays in CoT. NMR-Bench gains show the value of this mechanism for sustained instance discrimination and spatial continuity. The controlled interventions provide a more specific test: replacing inserted features changes task outcomes and can redirect later visual selection despite fixed intervening words. Together, these results give native multimodal reasoning a concrete implementation in which language and native representations jointly carry reusable intermediate states.

Retrieval is a first step toward a richer visual scaffold. Our model reinserts existing encoder features but does not explicitly compare or compose them into new states. Future systems could compare and compose visual features, or integrate 3D geometry and audio into the same reasoning chain. Whether these operations improve generalization beyond the controlled tasks remains to be tested.

\subsection*{AI use statement}
We used AI tools to assist with theoretical framework development, experimental design and result interpretation, code implementation, manuscript writing and translation, and formatting.

\bibliographystyle{iclr2027_conference}
\bibliography{reference}

\clearpage
\appendix
\section*{Supplementary Material}
\input{supplementary}

\end{document}

%% file: supplementary.tex
The appendices document implementation and training (Appendix~\ref{app:scaffold-details}), benchmark construction and scoring (Appendix~\ref{app:benchmark-details}), comparison settings (Appendix~\ref{app:main-results}), statistical procedures (Appendix~\ref{app:statistics}), content interventions (Appendix~\ref{app:evidence}), evidence-cap settings (Appendix~\ref{app:budget}), and visual-access diagnostics (Appendix~\ref{app:burden}).
\section{Visual Scaffolding: Implementation and Training Details}
\label{app:scaffold-details}

\subsection{Constructing visual intermediate states}
\label{app:architecture}
The two task models share the same visual-scaffolding architecture on Qwen3-VL-4B-Thinking. Language states and merged visual features both have width 2560. The standard visual bank enters through the backbone's native prefill and DeepStack path. A second view is enlarged twofold with Lanczos resampling and processed by the same complete vision encoder and spatial merger. Its merged features form the Dense Bank $V_d$, while the original-view merged features form the standard visual bank $V_s$. The shared vision encoder and spatial merger together are the Visual Encoder. The enlarged view's intermediate DeepStack outputs are not injected separately into the language layers. The processor supplies the dimensions of both grids.

To condition each visual selection on the recent reasoning state, the Visual Retriever uses eight-head context aggregation over at most 256 preceding positions. These can contain text, visual evidence, boundary markers, and initial prefill states. The current request supplies the query and is excluded from that context. The decoder retains its complete causal key--value history.

To encode local relations without changing which source cells are selectable, the spatial encoder preserves grid resolution. It applies a $1\times1$ channel projection from 2560 to 512, GroupNorm and GELU, two residual blocks each containing two $3\times3$ convolutions with GroupNorm and GELU, and a $1\times1$ projection back to 2560. The result is added to the original bank features and layer-normalized. Query and key projections each use three linear layers of widths $2560\to2560\to2560\to2048$, with LayerNorm and GELU after the first two layers. Sixteen 128-dimensional scoring heads are combined by learned, globally shared softmax weights. Sigmoid is applied independently over candidate cells.

Selected values come from the original dense-bank features. The Evidence Adaptor applies Equation~\ref{eq:scaffold-adapter} to these values, creating the visual intermediate states used by subsequent decoder steps. Both linear layers have width 2560. Initializing $W_2$ and $b_2$ to zero makes the initial Evidence Adaptor an identity map. The Visual Retriever and Evidence Adaptor contain 75,033,616 and 13,117,440 parameters, respectively, totaling 88,151,056 added parameters. Counting shared input/output embeddings once, the backbone has 4,437,815,808 parameters including vision, and the complete model has 4,525,966,864. The interface increases the parameter count by 1.986\%. All groups are finetuned.

\subsection{Training-target construction}
\label{app:training-targets}
Each example specifies a question, answer, reasoning trace, request positions, and GT patch sets. During training, the forward pass fills annotated evidence positions with adapted GT features. This lets the decoder learn a continuation from the intended visual support. Predicted retrieval scores are supervised at each request state: they determine the retrieval loss, while GT sets determine the inserted features. Teacher forcing applies to the preceding text and evidence, and both losses propagate through the differentiable computation.

Section~\ref{sec:scaffold-learning} defines the next-token, focal, and Dice objectives and their weights.

\subsection{Evidence budgets and sequence positions}
The actual special tokens corresponding to the paper's \texttt{[ref]} and \texttt{[end]} notation are \verb!<|img_ref|>! and \verb!<|ref_end|>!. The token \verb!<|ref_pad|>! reserves token slots for continuous evidence embeddings. When a request is generated, the selected evidence and its closing marker are supplied structurally rather than decoded one token at a time.

One reference event comprises a request and its evidence block, the unit manipulated in the content-intervention experiments. Training capacity is 64 evidence tokens per request. Ground-truth sets are retained as supplied, and exceeding this capacity is an error rather than a truncation rule. The training sets have maximum GT evidence-set sizes of 25 for Path and 18 for Counting. At inference, threshold 0.5 is applied first, top-$K$ caps an oversized selection, and a single argmax cell is used if the thresholded set is empty. Indices are sorted in row-major order before injection. Main comparisons use $K=8$, and the capacity study evaluates $K\in\{8,16,32\}$ with fixed weights. Content and burden diagnostics use $K=32$ throughout. Consequently, their unmodified accuracies differ from the main comparison, although the checkpoints are identical.

An emitted reference costs $K_t+2$ tokens: its request, $K_t$ evidence embeddings, and its closing marker. All committed generated tokens count toward the 16,384-token budget, which excludes the initial prefill. A reference is rejected if its complete span does not fit. Its evidence count is not shortened to consume the remaining budget. The engine processes the evidence and closing marker as one block and then resumes ordinary decoding. Subsequent tokens receive their normal sequence positions. The initial image retains the backbone's native multimodal positional encoding.

\subsection{Interface learning and joint adaptation}
\label{app:optimization}
Path and Counting use independent 40K and 32K training sets. Phase I covers one epoch and prioritizes the new interface with small, nonzero language and vision learning rates. Phase II increases backbone adaptation over the remaining epochs. Only final checkpoints are evaluated: epoch 4 (step 5000) for Path and epoch 7 (step 7000) for Counting. Table~\ref{tab:scaffold-training} summarizes the optimization settings, including the AdamW optimizer \citep{loshchilov2019adamw}.

\begin{table}[!htbp]
\centering\small
\caption{Optimization settings for the final models. Learning-rate triplets list the scaffold, language, and vision groups in that order.}
\label{tab:scaffold-training}
\begin{tabular}{@{}lll@{}}
\toprule
Setting & Path & Counting \\
\midrule
Training examples / epochs & 40,000 / 4 & 32,000 / 7 \\
Updates / Phase I updates & 5,000 / 1,250 & 7,000 / 1,000 \\
Global batch / microbatch per GPU & 32 / 1 & 32 / 1 \\
GPUs / gradient accumulation & 8 / 4 & 8 / 4 \\
Phase I peak / constant rates & $10^{-4},10^{-7},10^{-8}$ & same \\
Phase II peak rates & $5\!\times\!10^{-5},10^{-5},5\!\times\!10^{-6}$ & same \\
Phase I/II warmup & 5\% / 7\% & same \\
Phase II cosine minimum & 10\% of peak & same \\
Optimizer / weight decay & AdamW / 0.01 & same \\
Adam betas / epsilon & $(0.9,0.999)$ / $10^{-8}$ & same \\
Gradient clipping / training seed & 1 / 42 & same \\
Precision / maximum training length & BF16 / 12,800 & same \\
\bottomrule
\end{tabular}
\end{table}

Runs use eight A100 80GB GPUs, ZeRO-1, and gradient checkpointing. In Phase I, the scaffold learning rate warms up and decays to zero, while language and vision rates remain constant. Phase II uses warmup followed by cosine decay to 10\% of the peak. All training examples satisfy the sequence-length limit, with maximum lengths 6035 for Path and 2102 for Counting.

\section{NMR-Bench: Data and Scoring Details}
\label{app:benchmark-details}

\begin{table}[!htbp]
\centering
\caption{NMR-Bench training and test sets. Training examples include reasoning traces and patch annotations. Test inputs contain only an image and a question.}
\label{tab:nmr-data}
\small
\begin{tabular}{@{}lrrl@{}}
\toprule
Task & Training size & Test & Test stratification \\
\midrule
Burden Counting & 32,000 & 420 & 7 forms $\times$ 3 levels $\times$ 20 \\
Path Tracing & 40,000 & 360 & 4 marker types $\times$ 3 levels $\times$ 30 \\
\bottomrule
\end{tabular}
\end{table}

\subsection{Standard evaluation sets}
Path Tracing tests the maintenance of spatial continuity using 360 images on 512-, 768-, and 1024-pixel square canvases. It has 120 images at each level and 90 for each endpoint-marker type: letters, numbers, icons, and shapes. Curve counts are 3 for Easy, 4--5 for Medium, and 6--7 for Hard. Observed crossing counts are 0--5, 1--10, and 3--17, respectively. Curve count, layout, and visual interactions vary together in this standard set. Controlled variations are studied in Section~\ref{sec:exp_burden}.

Burden Counting tests instance discrimination through 420 images, balanced at 20 per task-form/difficulty cell. Its actual object totals range from 9 to 179. Canvas sizes are 512, 576, 640, 704, 768, and 832 pixels. The seven forms are listed below. Population ranges overlap across difficulty levels because attribute conditions and task forms impose different demands.

\begin{table}[!htbp]
\centering\small
\caption{Burden Counting task forms and canonical answer values.}
\label{tab:counting-forms}
\begin{tabular}{@{}lll@{}}
\toprule
Form & Query condition & Answer value \\
\midrule
T1a & Object category & Integer count \\
T2a & Category and size & Integer count \\
T2b & Category and tilt & Integer count \\
T2c & Category and occlusion & Integer count \\
T2d & Category, size, and tilt & Integer count \\
T3a & Compare large and small groups & Group label and count difference \\
T3b & Most frequent listed category & Category label and count \\
\bottomrule
\end{tabular}
\end{table}

\subsection{Training sets and annotation access}
The 40,000-example Path training set contains 6,000/12,000/22,000 Easy/Medium/Hard examples, 10,000 of each endpoint type, and approximately equal counts at the three canvas sizes. The 32,000-example Counting training set contains 7,000/14,000/11,000 Easy/Medium/Hard examples. T1a and T2a--T2c contain 4,800 examples each, T2d contains 2,829, and T3a/T3b contain 4,985/4,986.

Reasoning traces and patch targets provide training supervision. During evaluation, only the image and question enter generation. Answers and annotation metadata remain with the scorer. Standard test files do not provide the model with intermediate traces or selected regions. Spatial annotations used in the controlled diagnostic extensions are retained for offline analysis. Training and standard-test identifiers are disjoint. The standard evaluation sets also inform inference-capacity selection.

\subsection{Scoring task completion and partial progress}
To assess task completion, Counting answers are parsed into integers or the structured values in Table~\ref{tab:counting-forms}. Parsing first uses the final boxed answer and then the implemented fallback rules on the reasoning text. Exact accuracy requires the complete canonical value to agree with the stored target. MAE uses successfully parsed T1a and T2a--T2d answers, excluding the two comparison forms. Its denominator, at most 300, accompanies every reported value.

For the single tied T3a example (ten objects per group), scoring uses the stored canonical answer \texttt{small, 0}. The same rule applies to every model.

Path scoring normalizes label case, numeric forms, and escapes. Connections are treated as unordered pairs, and the complete answer as a set of such pairs. Duplicate pairs do not change that set. Exact accuracy requires equality with the ground-truth pair set. Unparseable predictions or predictions containing unknown labels receive zero. CRA follows Equation~\ref{eq:bench-metrics}, with invalid cases contributing zero. A parseable but incomplete answer can therefore receive partial CRA while failing exact accuracy. All invalid outputs remain in the accuracy denominator. Scoring is deterministic and uses no model-based judge.

\section{System Comparisons: Configurations and Scoring}
\label{app:main-results}

\subsection{Comparing visual reasoning interfaces}
The main table compares visual scaffolding with general-purpose MLLMs, coordinate-mediated interfaces, visual tools, and Qwen3-VL-4B-Thinking controls (Table~\ref{tab:nmr-bench-main}). The six general-purpose models are documented by their providers \citep{minimax_m3,claude_opus48,kimi_k26,qwen37_plus,gpt55,gemini31_pro}. The standard sets contain 420 Burden Counting and 360 Path Tracing examples, with results reported overall and by difficulty. Visual scaffolding uses the epoch-7 Counting and epoch-4 Path checkpoints with $K=8$. Table~\ref{tab:baseline-provenance} lists API model identifiers and length-limit stops. All examples remain in the accuracy denominator, including length-limited and unparseable outputs. The general-purpose baselines differ from our task-trained models in pretraining data, scale, and adaptation.

\begin{table}[!htbp]
\centering\footnotesize\setlength{\tabcolsep}{3pt}
\caption{API model identifiers and length-limit stops for both tasks. Stopped predictions remain in the accuracy denominator and are scored whenever they contain a parseable answer.}
\label{tab:baseline-provenance}
\begin{tabular*}{\linewidth}{@{\extracolsep{\fill}}llrr@{}}
\toprule
Table label & Requested model identifier & Counting & Path\\
\midrule
MiniMax-M3 & \texttt{MiniMax-M3} & 0 & 34\\
Claude-Opus-4.8 & \texttt{claude-opus-4-8} & 0 & 0\\
Kimi-K2.6 & \texttt{kimi-k2.6} & 197 & 75\\
Qwen-3.7-Plus & \texttt{qwen3.7-plus} & 40 & 34\\
GPT-5.5 & \texttt{gpt-5.5-ca} & 0 & 0\\
Gemini-3.1-Pro & \texttt{gemini-3.1-pro-preview} & 1 & 77\\
\bottomrule
\end{tabular*}
\end{table}

No-Ref uses the same task training data as visual scaffolding. It retains the \texttt{[ref]} trigger in the reasoning trace but omits visual-feature retrieval and insertion. The original image input remains available, while generation continues without newly inserted visual evidence. No-Ref answers 31/420 Counting and 191/360 Path examples correctly. Thus $\Delta_{\mathrm{NoRef}}$ measures the gain from adding visual scaffolding to the trigger-based reasoning process. Section~\ref{sec:exp_causal} separately tests the role of selected visual content within fixed model weights.

Counting MAE is evaluated on parseable integer-answer predictions, excluding the two comparison task forms. The valid denominators are 300 for MiniMax, Claude, Qwen-3.7-Plus, and GPT-5.5, 294 for Kimi, 298 for Gemini, 299 for Qwen3-VL-4B-Thinking, 295 for No-Ref, and 300 for visual scaffolding. For visual scaffolding, each difficulty contributes 100 integer-answer predictions. Path CRA averages the fraction of ground-truth connections recovered per image, assigning zero to unparseable predictions or answers containing unknown labels. The inference-capacity study evaluates the operating point on these standard test sets.

\section{Shared Statistical Protocol}
\label{app:statistics}
Uncertainty estimates follow the unit that varies in each comparison. Image-level content interventions and inference-cap comparisons use paired bootstrap 95\% confidence intervals. Content interventions stratify Path images by difficulty and Counting images by task-form--difficulty cell. Controlled image variants resample scene families, keeping related conditions together. All intervals use 10,000 bootstrap resamples and quantify variation across evaluation images, not independently trained models. Each experiment reports its cohort and denominators. Repeated anchors and donor draws are not counted as independent images.

\section{Testing the Functional Role of Visual Intermediate States}
\label{app:evidence}

\subsection{Separating selected content from the reference interface}
The content interventions ask whether the selected visual states do more than mark a retrieval action. All conditions use the final checkpoints documented in Appendix~\ref{app:optimization}, with weights fixed. Only content supplied through selected evidence positions is altered. Retrieval uses the 2$\times$ image branch, context-conditioned retrieval, a threshold of 0.5, a maximum of 32 selected features, and the Evidence Adaptor $\phi(v)=v+\mathrm{MLP}(v)$. Evidence tokens retain their ordinary increasing sequence positions. No module is retrained or removed.

The standard evaluation sets contain 120 Path examples per difficulty level and 20 Counting examples per task-form--difficulty cell (seven forms, three levels). Trajectory-level interventions measure the cumulative role of selected content, while local branches isolate a change at a shared reasoning prefix. The local cohort contains 32 Path examples per difficulty (96 total) and four Counting examples per form--difficulty cell (84 total), sampled with seed 20260910 independently of correctness. All 780 examples appear in each global condition. We denote unmodified inference by Normal.

For a selected index set $S$ of size $K_t$, Normal writes $\phi(V_d[S])$. Zero replaces the \emph{output} of $\phi$ with zeros, leaving the structural end-token embedding unchanged. Mismatch samples $K_t$ distinct donor indices $D$ uniformly from the same reference grid outside $S$ and uses $\phi(V_d[D])$. A local random generator is derived from the seed, sample identifier, and reference index. It does not consume decoding RNG. Donors and a random slot ordering are shared between the 50\% and 100\% conditions at a common prefix. The half condition replaces $\lfloor K_t/2\rfloor$ slots, plus one with probability $1/2$ when $K_t$ is odd. The realized replaced fraction is recorded. Donors change the selected image locations while retaining the same-image feature distribution and its contextual information.

Greedy generation uses a common 16,384-token budget, including ordinary generated tokens and every committed reference action. A complete action costs $K_t+2$ tokens: trigger, $K_t$ evidence slots, and end token. An action that does not fit is not partially committed. The original image prefill, visual features, and retrieval-key cache remain available and unmodified. Each intervention preserves the current action's $K_t$ and positions, but subsequent free trajectories may differ in wording, reference count, selected $K_t$, and length. Zero is therefore a content intervention in the reference-enabled model, not a No-Ref baseline.

\subsection{Changing one intermediate visual state}
For a Normal trajectory with $R$ references, local anchors are $\lceil0.25R\rceil$, $\lceil0.5R\rceil$, and $\lceil0.75R\rceil$ (one-based). All sampled trajectories admit three distinct anchors. Each branch replays the model's own Normal prefix through one anchor trigger with the same context aggregation and retrieval computations, checks the predicted indices and token positions, and then resumes unconstrained generation. No ground-truth text or reference indices are supplied. Independent generation calls rebuild mutable KV and hidden-state buffers. Prefix tokens still count toward the common budget. Unmodified sham branches reproduce Normal exactly.

The local intervention replaces only the chosen reference, with later references operating normally. It includes three time positions, two additional donor seeds at the middle position (20260911 and 20260912), and a middle-reference control that rescales each donor vector to the original slot's L2 norm before injection. A matched \emph{half-reference} middle control uses the same 180-example cohort, with seed 20260910 and the same donor sets and slot ordering as complete replacement. Pairing the prefix, donors, and slots allows the replacement fraction to vary at the same reasoning state. Its mean realized replaced fractions are 49.84\% for Path and 49.98\% for Counting.

\begin{table}[!htbp]
\centering
\footnotesize
\setlength{\tabcolsep}{3pt}
\caption{\textbf{Paired local effects relative to sham.} Values are percentage-point changes with 95\% confidence intervals. The 96 Path and 84 Counting images have sham accuracies of 60.42\% and 58.33\%, respectively. ``Half'' uses the same donors and prefix as complete replacement.}
\label{tab:evidence_local}
\begin{tabular}{@{}lrrr@{}}
\toprule
Intervention & Path $\Delta$Acc. & Path $\Delta$CRA & Counting $\Delta$Acc. \\
\midrule
Early, all & $-41.7\;[-52.1,-31.2]$ & $-30.7\;[-37.1,-24.3]$ & $-31.0\;[-41.7,-20.2]$ \\
Middle, all & $-41.7\;[-52.1,-32.3]$ & $-30.2\;[-36.1,-24.4]$ & $-22.6\;[-34.5,-10.7]$ \\
Late, all & $-21.9\;[-32.3,-11.5]$ & $-16.1\;[-22.1,-10.4]$ & $-22.6\;[-33.3,-11.9]$ \\
Middle, seed 2 & $-38.5\;[-49.0,-28.1]$ & $-27.5\;[-34.2,-21.1]$ & $-20.2\;[-32.1,-8.3]$ \\
Middle, seed 3 & $-42.7\;[-52.1,-33.3]$ & $-31.2\;[-37.5,-24.9]$ & $-32.1\;[-42.9,-21.4]$ \\
Middle, norm matched & $-41.7\;[-52.1,-31.2]$ & $-26.3\;[-31.9,-20.9]$ & $-22.6\;[-34.5,-10.7]$ \\
Middle, half & $-7.3\;[-14.6,0.0]$ & $-8.1\;[-13.0,-3.6]$ & $+2.4\;[-2.4,7.1]$ \\
\bottomrule
\end{tabular}
\end{table}

Table~\ref{tab:evidence_local} reports local effects at matched prefixes and across positions in the unmodified trajectory. On Path, early complete replacement is more damaging than late replacement by 19.79 accuracy points (paired 95\% CI $[-31.25,-8.33]$ for early minus late). The CRA contrast is $-14.62$ points ($[-21.84,-7.29]$). The Counting accuracy contrast is $-8.33$ points, with an interval spanning zero ($[-20.24,2.38]$). These position comparisons concern different states in the model's own trajectory, each with a different remaining reasoning horizon.

Matching feature norms preserves the aggregate accuracy effects of middle-reference replacement. The trajectories can still differ: complete token sequences match between unscaled and norm-matched branches in 5/96 Path and 23/84 Counting cases. Both extra donor seeds also retain substantial negative effects. The effect persists across feature-norm matching and donor choices, strengthening the attribution to selected visual content.

\subsection{Subsequent visual selection under fixed intervening text}
We distinguish an immediate language-model response from the next visual decision. The immediate-response diagnostic computes full-vocabulary Jensen--Shannon divergence and argmax change at the post-reference decision, including action and termination tokens. For the middle-reference next-query probe, both branches receive the same Normal-generated intervening text up to the next reference trigger, and execute the Visual Retriever with its context aggregation. Under the shared text, both branches predict their own index sets and retrieval scores, which we then compare.

The next selected set changes in 71/96 Path examples (73.96\%, 95\% CI $[65.63,82.29]$) and 55/84 Counting examples (65.48\%, $[55.95,75.00]$). Mean Jaccard similarities between selected sets are 0.363 and 0.449. Selected cardinality changes by $+7.04$ and $+0.83$ features on average. Because the intervening token sequence is shared, these differences identify an effect of earlier visual content on the next visual decision under fixed local wording. The effect includes information propagated through subsequent hidden states.

\begin{figure}[!t]
\centering
\includegraphics[width=.88\linewidth]{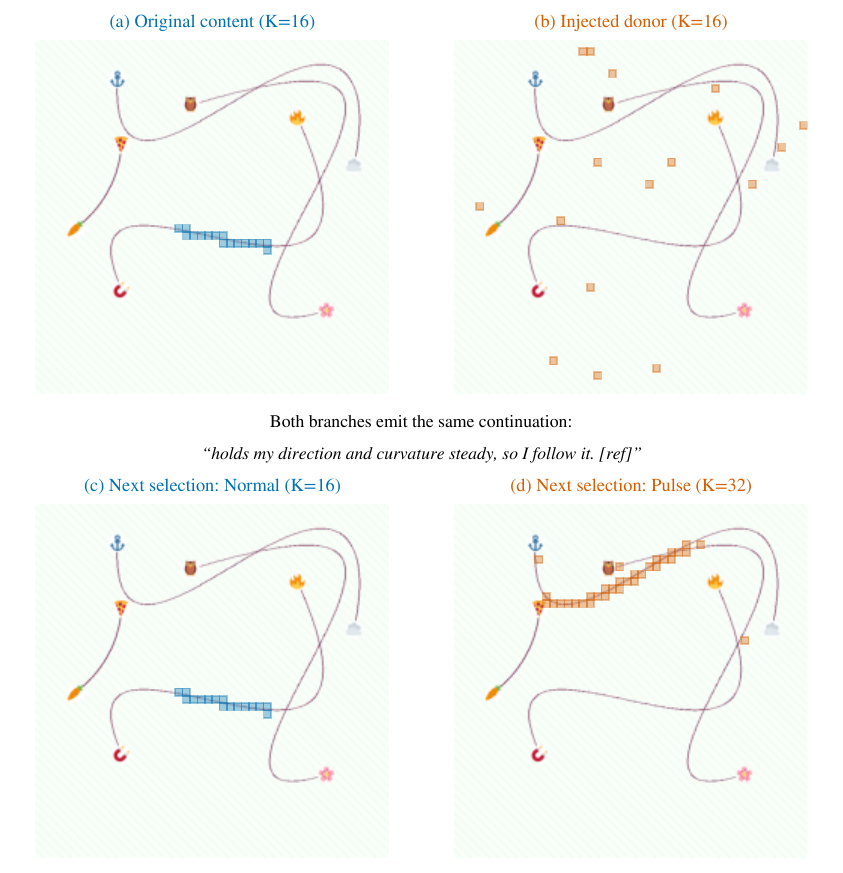}
\caption{\textbf{Identical words, different visual decisions.} (a) Selected content and (b) same-image replacement. Both branches emit the displayed text, but their next selected regions (c,d) are disjoint. Normal recovers all four pairs, while replacement recovers two. The case is closest to median immediate JSD ($2.97\times10^{-5}$ nats) among the 13 correct-to-incorrect Path cases with identical nonempty intervening text and changed selections. Overlays show source cells. Panel labels report actual block sizes, subject to a cap of 32.}
\label{fig:evidence_same_text}
\end{figure}

Under free continuation, 36 Path branches preserve the same emitted sequence through the next trigger. Of these, 33 change the next selected region, and 13 change a correct Normal answer to an incorrect one. All 33 include nonempty identical intervening text. Counting has 45/84 cases with the same emitted sequence and a different next selection, but 44 are consecutive reference actions with no intervening words. We distinguish this action-level identity from the nonempty verbal identity illustrated for Path in Figure~\ref{fig:evidence_same_text}. More broadly, 31 of the 42 harmed Path middle-reference examples and 16 of the 29 harmed Counting examples retain the same immediate next-token argmax. Thus, selected visual content can affect later reasoning while preserving the immediate next token.

\FloatBarrier

\section{Inference Capacity and Evidence Use}
\label{app:budget}

To evaluate the inference operating point in Section~\ref{sec:exp_reference_policy}, we use both final checkpoints on all 780 standard test images at evidence caps $K\in\{8,16,32\}$. Thresholding precedes the cap, so $K$ is a maximum rather than a fixed number per request. Model weights and decoding settings are held fixed across the sweep. The training interface has capacity for 64 annotated features per request.

\begin{table}[!htbp]
\centering\small
\caption{Mean evidence tokens per image on the standard test sets. A smaller cap reduces the visual material inserted into the reasoning trace.}
\label{tab:budget-cost}
\begin{tabular}{@{}lrrr@{}}
\toprule
Task & $K=8$ & $K=16$ & $K=32$\\
\midrule
Burden Counting & 171.8 & 201.5 & 204.3\\
Path Tracing & 346.6 & 528.3 & 764.9\\
\bottomrule
\end{tabular}
\end{table}

The fixed-weight accuracy gains at $K=8$ relative to $K=32$ are 10.95 percentage points for Counting (95\% CI $[6.43,15.48]$) and 13.06 points for Path ($[8.61,17.50]$). Each paired comparison resamples test images and keeps both cap predictions for an image together.

\section{Visual Access Across Increasing Demands}
\label{app:burden}

This appendix documents the controlled conditions plotted in Figure~\ref{fig:capacity-access}(b). Both frozen task models use greedy decoding, a retrieval threshold of 0.5, $K=32$, and the same 16,384-token generation budget as the standard evaluation. Ground-truth masks are used only after inference to measure which source-image regions were selected. Coverage is a spatial-access measure; it does not score answer correctness.

\subsection{Counting: fixed-total target variation}
The Counting control uses the ordinary category-counting task (T1a) on $832\times832$ canvases. Each of 24 scene families has 128 fixed object locations. Target identities are varied to create 32, 64, or 96 targets, while distractor identities fill the remaining locations. The layouts and target-category assignments are nested within each family. Icons are 32 pixels wide, have zero rotation, and do not overlap in alpha support. The model and inference settings are unchanged across conditions.

\subsection{Path: fixed-curve crossing variation}
Path images use $768\times768$ canvases, six curves, 3-pixel strokes, white backgrounds, and uniform curve color. Each crossing condition contains 32 geometries balanced across letter and two-digit endpoint labels and sparse and mixed layouts. Exact crossing counts are 3, 6, 9, and 12. Crossings are computed between sampled polylines with a one-pixel deduplication rule. The textual query format and rendering protocol are fixed across conditions.

\subsection{Coverage calculation}
Selected patch indices are mapped through the merged reference grid to their source-image regions. If $U$ is the union of regions selected over a complete trace and $M_j$ is the visible mask of target object $j$, Counting coverage is the per-image average of $|M_j\cap U|/|M_j|$ over targets. Path coverage is $|M\cap U|/|M|$, where $M$ is the union of visible curve pixels outside endpoint circles. The reported percentages average these image-level values within each condition. The corresponding values appear in Figure~\ref{fig:capacity-access}(b).

Uncertainty intervals in Figure~\ref{fig:capacity-access}(b) use 10,000 bootstrap resamples of scene families for Counting and geometries for Path. Related images within a family remain together in a resample.